\documentclass[conference]{IEEEtran}
\IEEEoverridecommandlockouts
\usepackage{cite}
\usepackage{amsmath,amssymb,amsfonts}
\usepackage{algorithmic}
\usepackage{graphicx}
\usepackage{textcomp}
\usepackage{xcolor}
\usepackage{booktabs}
\usepackage{tabularx}
\usepackage{bm}

\usepackage{graphics} 
\usepackage{amsmath} 
\usepackage{amssymb}  
\usepackage{algorithm}
\usepackage{algorithmic}
\usepackage{xcolor}
\usepackage{enumitem}

\newcommand{\ba}{\begin{array}}
	\newcommand{\ea}{\end{array}}
\newcommand{\babc}{\begin{abc}}
	\newcommand{\eabc}{\end{abc}}
\newcommand{\bc}{\begin{center}}
	\newcommand{\ec}{\end{center}}
\newcommand{\be}{\begin{equation}}
\newcommand{\ee}{\end{equation}}
\newcommand{\bea}{\begin{eqnarray}}
\newcommand{\eea}{\end{eqnarray}}
\newcommand{\beas}{\begin{eqnarray*}}
	\newcommand{\eeas}{\end{eqnarray*}}
\newcommand{\bh}{\begin{hangitem}}
	\newcommand{\eh}{\end{hangitem}}
\newcommand{\bhi}{\begin{hangitem}}
	\newcommand{\ehi}{\end{hangitem}}
\newcommand{\bn}{\begin{enumerate}}
	\newcommand{\en}{\end{enumerate}}
\newcommand{\bq}{\begin{quote}}
	\newcommand{\eq}{\end{quote}}
\newcommand{\btb}{\begin{tabular}}
	\newcommand{\etb}{\end{tabular}}
\definecolor{cmt}{RGB}{255, 0, 0}

\def\litem[#1]{\item[#1\hfill]}         
\usepackage{csquotes}
\usepackage{multirow}
\usepackage{multicol}
\usepackage{cite}
\usepackage{float}
\usepackage{array}
\newcolumntype{L}[1]{>{\raggedright\let\newline\\\arraybackslash\hspace{0pt}}m{#1}}
\newcolumntype{C}[1]{>{\centering\let\newline\\\arraybackslash\hspace{0pt}}m{#1}}
\newcolumntype{R}[1]{>{\raggedleft\let\newline\\\arraybackslash\hspace{0pt}}m{#1}}
\usepackage{color}

\usepackage{graphicx}
\def\BibTeX{{\rm B\kern-.05em{\sc i\kern-.025em b}\kern-.08em
    T\kern-.1667em\lower.7ex\hbox{E}\kern-.125emX}}
\begin{document}

\title{Prospect Theory in Physical Human-Robot Interaction: a Pilot Study of Probability Perception\\
}

\author{\author{Yixiang Lin, Tiancheng Yang, Jonathan Eden and Ying Tan 
\thanks{$^{1}$Yixiang Lin, Tiancheng Yang, Jonathan Eden and Ying Tan are with the Department of Mechanical Engineering, The University of Melbourne, 3010 Parkville VICTORIA, Australia
{\tt\small yixiang.lin.1@student.unimelb.edu.au, tiancheng.yang1@student.unimelb.edu.au, eden.j@unimelb.edu.au, yingt@unimelb.edu.au}}%
}

}

\maketitle


\begin{abstract}
Understanding how humans respond to uncertainty is critical for designing safe and effective physical human–robot interaction (pHRI), as physically working with robots introduces multiple sources of uncertainty, including trust, comfort, and perceived safety. Conventional pHRI control frameworks typically build on optimal control theory, which assumes that human actions minimize a cost function; however, human behavior under uncertainty often departs from such optimal patterns. To address this gap, additional understanding of human behavior under uncertainty is needed. This pilot study implemented a physically coupled target-reaching task in which the robot delivered assistance or disturbances with systematically varied probabilities ($10\%$ to $90\%$). Analysis of participants' force inputs and decision-making strategies revealed two distinct behavioral clusters: a "trade-off" group that modulated their physical responses according to disturbance likelihood, and an "always-compensate" group characterized by strong risk aversion irrespective of probability. These findings provide empirical evidence that human decision-making in pHRI is highly individualized and that the perception of probability can differ to its true value. Accordingly, the study highlights the need for more interpretable behavioral models, such as cumulative prospect theory (CPT), to more accurately capture these behaviors and inform the design of future adaptive robot controllers.
\end{abstract}

\begin{IEEEkeywords}
Physical Human-Robot Interaction, Cumulative Prospect Theory,  Human Behavior Modeling.
\end{IEEEkeywords}

\section{Introduction}

Physical human–robot interaction (pHRI) refers to human–robot interactions that involve physical contact or force exchange \cite{haddadin2016physical, farajtabar2024path}. PHRI enables humans and robots to share workspaces and perform collaborative tasks such as assembly, object handover, co-manipulation of tools, and assisted rehabilitation \cite{ajoudani2018progress, matheson2019human, mohebbi2020human}. Unlike conventional industrial robots that operate within safety barriers or cages, pHRI occurs in close proximity to humans, where physical coupling means that the robot’s actions can directly influence human movements and decisions \cite{losey2018review}. Consequently, robots in pHRI must account for factors critical to safe and effective collaboration, including real-time decision-making, shared control between agents, and mutual adaptation \cite{lasota2017survey, losey2018review}.

Current robot control approaches in pHRI often rely on optimal control or game-theoretic frameworks, which assume that interacting agents seek to optimize performance metrics such as efficiency, robustness, or energy expenditure \cite{li2019differential}. These approaches can incorporate variability in human behavior through stochastic optimal control \cite{berret2021stochastic} or noisy-rational models \cite{kwon2020humans}, and game-theoretic formulations can model interactions where multiple agents pursue coupled objectives \cite{li2019differential}. While these methods provide a rigorous mathematical foundation for robot decision-making, they generally assume that humans behave near-optimally, minimizing or maximizing a predefined cost or utility function.

However, this assumption often fails in high-uncertainty or high-risk scenarios. Human behavior under uncertainty frequently deviates from optimal patterns due to factors such as limited information, risk perception, prior experience, and individual biases \cite{ota2015motor, wolpert2012motor}. In pHRI, unpredictable task dynamics and incomplete information about the robot’s sensory feedback, intentions, or future movements create substantial uncertainty for the human partner \cite{losey2018review}. Moreover, individuals perceive and respond to this uncertainty in highly personalized ways: they assign different weights to perceived gains and losses and interpret the probabilities of these outcomes differently. These individual differences shape human behavior during collaboration and, in turn, influence what constitutes an effective and adaptive robot controller.

To better understand and model these non-optimal human behaviors, neuroscientists and behavioral scientists have proposed alternate theories that account for the non-rational aspects of human decision-making \cite{ho2022cognitive}. Among these theories, cumulative prospect theory (CPT) \cite{tversky1992advances} has proven particularly effective for modeling decision-making under uncertainty and risk. Originally developed in economics and behavioral sciences, CPT accounts for non-rational, individualized behavior by incorporating reference points, loss aversion, and nonlinear weighting of probability \cite{kahneman1979prospect, barberis2013thirty, gurevich2009decision}. Within human–robot interaction (HRI), CPT can capture phenomena such as asymmetric trust, algorithm aversion, and framing effects \cite{de2020predicting, smith2025if}, where humans weigh potential negative outcomes more heavily than equivalent positive outcomes. This can provide a principled approach for designing robot controllers that can adapt to individual human behaviors, supporting safer and more effective collaboration.

While CPT has been applied in human behavior modeling \cite{tian2021bounded} and human–robot interaction\cite{kwon2020humans}, it has not yet been extended to scenarios involving physical interaction. Existing CPT frameworks do not fully consider the implicit factors that naturally arise during physical collaboration, such as effort, comfort, trust, and safety \cite{gervasi2022user, ogenyi2019physical}. How these factors interact with personalized uncertainty to shape human decisions—-whether to accept, reject, or adapt to robotic actions-—remains poorly understood \cite{hancock2021evolving}. To begin to address this gap, we designed a controlled, physically coupled pilot target-reaching experiment in which participants physically interacted with a robot under probabilistic disturbances. By systematically varying the probability of robot actions and recording participants’ choices and behavior, the experiment presents an initial evaluation of how humans respond to uncertainty, effort, and potential losses during real-time physical human-robot collaboration. This data and its analysis can inform future application of CPT to model human decision-making in pHRI, offering actionable insights for designing safer, more adaptive, and user-tailored collaborative robotic systems.


\section{Experiment}\label{sec_Experiment}


\subsection{Participants}

The experiment was approved by the University of Melbourne Low-and-Negligible-Risk (LNR) Ethics Committee (reference number: 2025-27757-71150-5). A total of 10 participants (8 males and 2 females; age 27.7$\pm$3.6 years old [mean$\pm$std]) from diverse backgrounds took part in the study. All participants were right-hand dominant and provided informed consent prior to commencing the experiment.
\subsection{Experimental Setup}

\begin{figure}[htb!]
\centering
\includegraphics[width=0.95\linewidth]{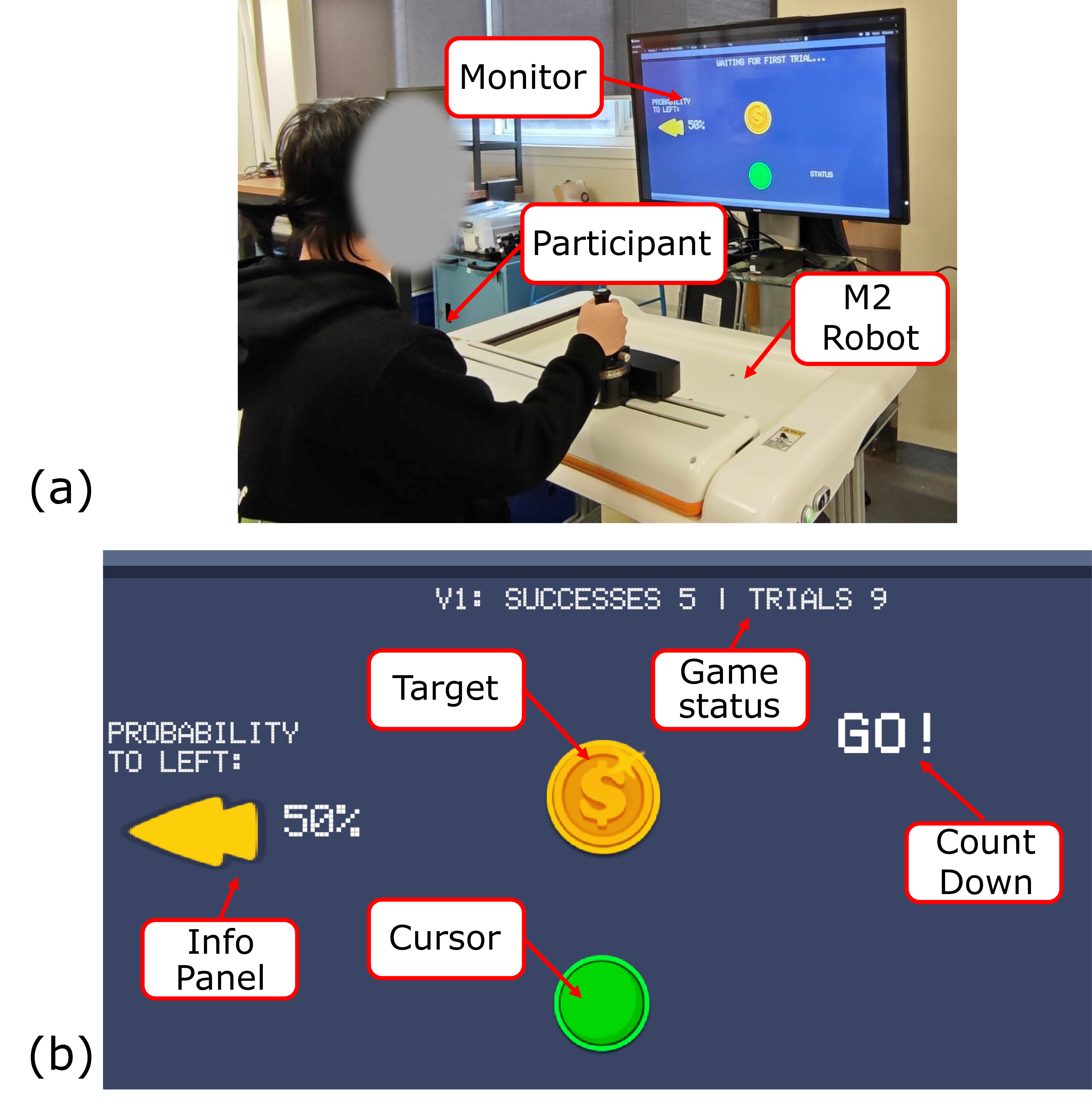}
\caption{Experimental setup. (a) Based on the information displayed on the monitor, the participant physically interacts with the ArmMotus M2 robot via the handle.
(b) The monitor presents the game status and target area. Information about the robot’s behavior is shown on the left information panel. The green cursor provides real-time feedback on the robot’s handle position. A “3, 2, 1, Go!” countdown signals the start of each trial.}
\label{Fig. 1}
\end{figure}

The experimental setup is shown in Fig. \ref{Fig. 1}. Participants interacted with the Fourier ArmMotus M2 robot (https://www.fftai.com/products-armmotus-m2-pro), an X–Y table haptic device that supports two-dimensional translational motion and records force, acceleration, and velocity along the X- and Y-axes. The robot’s handle served as a shared interface, with participants providing manual forces as input, while the robot applied forces via its motors. Participants received essential task-related information and real-time visual feedback of the handle position through a graphical user interface (GUI) displayed on a monitor positioned in front of the robot. The GUI was developed in Unity (https://unity.com), a real-time development platform widely used for interactive applications and video games. All device integration and communication were handled using ROS2 (https://www.ros.org).


\subsection{Experiment Task}
The experimental task involved a collaborative target-reaching activity between a human and a robot, performed under conditions of uncertainty. The robot could perform one of two possible robot actions (\textit{RA}):
\begin{itemize}
    \item[] \textbf{RA1 (UP)}: The robot moved directly forward toward the target. This represented an assistance trial.
    \item[] \textbf{RA2 (Side):} The robot provided a disturbance by moving sideways to the left.
\end{itemize}

Each trial began with the robot handle placed at a consistent origin position with the target reach area displayed 25\,cm directly in front of this position.  Here, participants were instructed to take one of two actions before the visualized countdown ended:
\begin{itemize}
\item[] \textbf{HA1 (Relax)}: The participant would let the robot assist in going to the target area without any preemptive effort. 
\item[] \textbf{HA2 (Compensate)}: The participant would actively compensate for the potential sideways disturbance by providing a counteracting rightward movement.
\end{itemize}

To inform participants about the likelihood of robot interference, the GUI displayed the probability that the robot would execute the disturbing action \textit{RA2} (referred to as the perturbation probability). 

At the end of the countdown, participants were instructed to start the reach with a ``Go!'' signal displayed on the monitor. They then had 0.5 seconds to complete the reach, a duration chosen to encourage primarily ballistic movements (i.e., fast, preplanned, feedforward movements executed without online sensory feedback corrections). A trial was considered successful only if the robot handle reached a circular target region with an $8$\,cm radius within this time limit. The trade-off, or ``gain versus loss'', in this task involved balancing the risk of conserving effort against the potential benefit of performing a feedforward compensation to reduce the likelihood of failure.

\subsection{Protocol}
\begin{figure*}[h!] 
\centering
\includegraphics[width=0.9\linewidth]{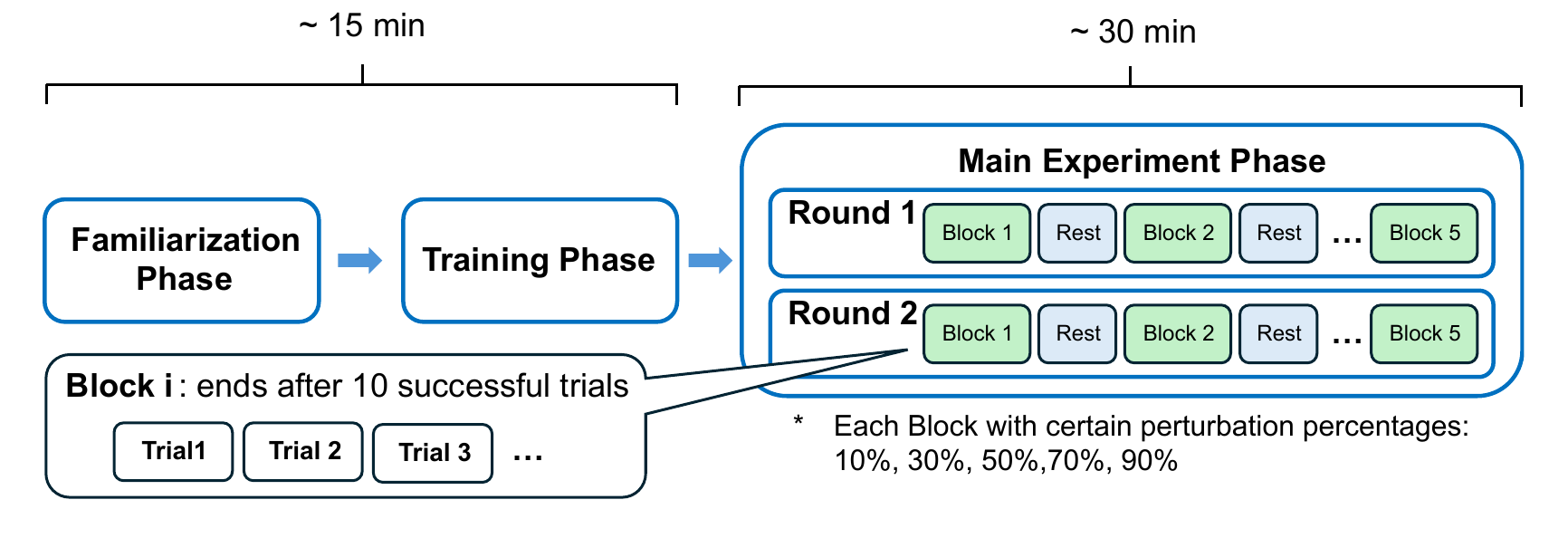}
\caption{The experiment protocol. The experiment was comprised of three phases: familiarization, training and main experiment. The main experiment phase was then itself divided into two rounds each containing 5 blocks for which all trials had a fixed perturbation probability. A single block was completed only after the participant successfully executed 10 reaches.}
\label{Fig. 2}
\end{figure*}

The experiment protocol is shown in Fig. \ref{Fig. 2}. The experiment consisted of three phases: \textit{familiarization}, \textit{training} and \textit{main experiment}. 

\noindent {\underline{Familiarization Phase}} 

During the \textit{familiarization phase}, participants were first briefed on the experiment and allowed to freely interact with the robot to become comfortable with the setup. Next, the magnitude of the disturbance for \textit{RA2} was calibrated to each participant's arm strength to ensure the task was challenging but manageable. To do this, participants applied a rightward force that was at the limit of what they could manageably hold for three seconds. The robot perturbation force was then set to 0.8 times this recorded value, keeping it slightly below the participant's maximum sustainable force. This calibration ensured that participants would need to exert effort—allowing the study to consider the cost of exertion—while minimizing the risk of cumulative muscle fatigue.

\noindent {\underline{Training Phase}} 

Once the disturbance force was calibrated, participants proceeded to the \textit{training phase}. In this phase, for each robot action (\textit{RA}, which includes \textit{RA1} and \textit{RA2}), participants completed a block of trials to demonstrate that they could successfully respond with the corresponding human action (\textit{HA}, which includes \textit{HA1} for \textit{RA1} and \textit{HA2} for \textit{RA2}). Each training block required participants to perform 10 successful reaches. Combined, the familiarization and training phases lasted approximately 15 minutes.

\noindent {\underline{Main Experiment Phase}} 

The \textit{main experiment phase} required participants to perform the same reaches as in the training phase, but with uncertainty regarding which robot action (\textit{RA}) would occur on a given trial. During this phase, each participant completed two rounds, each consisting of five blocks of reaches, with the entire phase lasting approximately 30 minutes. Each block corresponded to a series of reaches with RAs pre-generated using random seeds at fixed perturbation probability, denoted as $P_R=P(RA2)$, of $10\%$, $30\%$, $50\%$, $70\%$, and $90\%$. To minimize potential learning effects, participants were randomly assigned to two groups: one group completed the rounds in ascending order of perturbation probability ($10\%$ to $90\%$), while the other completed them in descending order ($90\%$ to $10\%$). As in the \textit{Training Phase}, participants were required to successfully execute 10 reaches to complete a block, with a reach considered successful if the robot handle reached the target region within the 0.5-second time limit.  A 30-second rest period was provided between blocks to reduce fatigue. 

Each reaching trial consisted of three components:
    \begin{enumerate}
        \item \textit{Ready:} Participants selected their intended HA and positioned themselves. For \textit{HA1}, participants relaxed; for \textit{HA2}, they applied a preload force to the handle. A virtual latch prevented premature movement. The countdown and “Go!” signal marked the end of this phase.
        \item \textit{Movement:} The virtual latch was released, allowing free handle movement for 0.5 seconds. The trial was evaluated as successful if the handle reached the target within this time.
        \item \textit{Reset:} The handle automatically returned to the start, the virtual latch was reactivated, and the system prepared for the next trial.
    \end{enumerate}


\subsection{Data Analysis}
\noindent {\underline{Data Processing}} 

Using the recorded force data, participants' actions were classified into the two instructed \textit{HAs}. Specifically, if a participant applied an average force greater than 10\,N for at least 1-second before the “Go!” signal, the trial was classified as a \textit{HA2} (compensate) action; otherwise, it was classified as a \textit{HA1} action. 

For each block of trials, the robot applied a constant probability of perturbing $P_R$ at one of the values of $10\%$, $30\%$, $50\%$, $70\%$, or $90\%$. The participant's tendency to compensate was quantified using the \textit{compensation probability}, $P_2$, representing the probability of choosing the compensation action (\textit{HA2}) for that $P_R$. This was calculated as
\bea
P_2 := P(\text{\textit{HA2}}) = \frac{N_{\text{\textit{HA2}}}}{N_{\text{total}}},
\label{eqn_P2}
\eea
where $N_{\text{\textit{HA2}}}$ is the number of trials classified as \textit{HA2} and $N_{\text{total}}$ is the total number of trials in the block. For example, if a participant performed \textit{HA2} in 7 out of 10 trials for $P_R = 30\%$, the compensation probability for that block would be 0.7. This provides a block-wise measure of participants' likelihood to compensate under different perturbation probabilities.


Two models were considered to explain the observed data for each participant. The first was a purely data-driven \textit{Bayesian Logistic Regression} (BLR) \cite{hilbe2016practical}, and the second was a model-based \textit{CPT} approach used to approximate the relationship between the participant's compensation probability, $P_2$, and the robot's perturbation probability, $P_R$. Here, when $P_2 = P_R$, it indicates that the participant's compensation probability matches the true probability of the robot perturbation.

The choice to use BLR was motivated by several factors  \cite{harris2021primer, johnson2022bayes}. First, the outcome of each trial is binary (\textit{HA1}  or \textit{HA2}), making logistic regression a natural model for the log-odds of the probability of compensation. Second, the Bayesian approach allows us to treat the regression parameters as random variables, providing a principled way to quantify uncertainty in the estimated relationship. Third, BLR is robust to small sample sizes and can incorporate prior information to stabilize estimates. Finally, the logistic function captures the non-linear, sigmoidal mapping between perturbation probability ($P_R$) and the likelihood of compensation ($P_2$), which aligns with observed human behavior under uncertainty.

CPT was selected to model participant behavior because it provides a principled framework for capturing the non-rational aspects of human decision-making under uncertainty. Unlike standard optimal control approaches that assume near-optimal behavior, CPT accounts for several key features observed in our task. First, CPT can model the fact that individuals perceive probabilities differently from the objective probabilities. For example, participants may overweigh rare events and underweigh common ones, leading to systematic deviations from the actual perturbation probability ($P_R$). This feature allows the model to capture how human perception of risk influences their choice of the compensation action (\textit{HA2}).
 Second, CPT is reference-dependent, evaluating gains and losses relative to individual expectations rather than absolute outcomes, which aligns with our observation of highly individualized participant responses. Third, it incorporates loss aversion, explaining why some participants may consistently compensate (\textit{HA2}) even at low perturbation probabilities. Finally, CPT allows its parameters, such as probability weighting and loss aversion, to vary across individuals, enabling the model to capture personalized patterns of risk perception and effort-reward trade-offs. Together, these features make CPT a suitable framework for modeling human behavior in our physically coupled target-reaching task under uncertainty.

\noindent {\underline{BLR-Based Modelling}} 

The BLR method incorporates Bayesian prior regularization into logistic regression and is well suited for modeling probabilities in binary classification problems. Within this framework, a participant’s compensation probability $P_2$ estimated using BLR, denoted as $P_2^{B}$, is given by
\begin{equation}
P_{2}^B = \frac{1}{1 + e^{-(\bar{\beta}_0 + \bar{\beta}_1 \cdot P_{R})}},
\label{eq 1}
\end{equation}
where $P_2^B$ is computed from experimental data as in (\ref{eqn_P2}). The intercept parameter $\bar{\beta}_0$ reflects the participant's baseline willingness to compensate, while the slope parameter $\bar{\beta}_1$ measures their sensitivity to changes in the robot's perturbation probability ($P_{R}$). A high positive $\bar{\beta}_0$ indicates a risk-averse strategy, as the participant maintains a high compensation likelihood regardless of the probabilistic cue. Similarly, a large positive $\bar{\beta}_1$ corresponds to a steep choice function approaching a step-like response, meaning the decision to compensate shifts abruptly within a narrow range of perceived risk. 

Parameter estimation for the BLR model was performed separately for each participant using their individual choice data. 
For each participant, a gradient-based maximum a posteriori (MAP) optimisation was first applied to obtain stable initial parameter values, 
followed by Hamiltonian Monte Carlo sampling with the No-U-Turn Sampler (NUTS) \cite{chen2014stochastic} to refine the estimates. 
Both BLR parameters were assigned weakly informative Gaussian priors with a standard deviation of 5 to ensure numerical stability during estimation. 
Because the BLR formulation does not impose an upper bound on the intercept parameter $\bar{\beta}_0$, 
logistic regression can produce extremely large or undefined estimates when a participant consistently chooses the same action. 
To avoid this issue, we capped $\bar{\beta}_0$ at 10 whenever the optimisation attempted to return an unbounded value.

\noindent {\underline{CPT-Based Modelling}} 

To better understand the decision-making mechanisms underlying these behavioral patterns, we modelled participant action choices using CPT \cite{tversky1992advances}, which accounts for the subjective evaluation of gains, losses, and probabilities.

Let $a_i$ denote a participant's action, where $a_1$ corresponds to selecting HA1 and $a_2$ corresponds to selecting HA2, and $RA_j$ denote a robot action, where $RA_1$ and $RA_2$ represent the two different robot perturbations used in the experiment. Additionally, let $\omega(a_i, RA_j)$ represent the subjective value of the outcome when the participant chooses $a_i$ and the robot executes $RA_j$.  Denoting the perceived reward for a successful trial be $V$, the perceived loss for a failed trial be $-G$, and the subjective effort cost be $C$ for a positive triplet $(V, G, C)$  Under this framework, the subjective value $\omega(\cdot,\cdot)$ for different combinations of participant and robot actions is summarised in Table~\ref{tab:reward_cost}.

\begin{table}[htb!]
\centering
\caption{Subjective value $\omega(a_i, RA_j)$ for each combination of participant action $a_i$ and robot action $RA_j$.}
\label{tab:reward_cost}
\begin{tabularx}{0.45\textwidth}{|>{\centering\arraybackslash}X|>{\centering\arraybackslash}X|>{\centering\arraybackslash}X|}
\hline
 & $a_1$ (HA1) & $a_2$ (HA2) \\
\hline
$RA_1$ & $V$ & $V-C$ \\
\hline
$RA_2$ & $-G$ & $V-C$ \\
\hline
\end{tabularx}
\end{table}

In CPT, the subjective value depends on the outcome (\(\omega(a_i, RA_j)\)), and probabilities are distorted through a weighting function $\pi(\cdot)$. The CPT-based utility of a given action $a_i$ is therefore
\begin{equation}
U^{\mathrm{C}}(a_i)
= \omega(a_i, RA_1) \cdot \pi(P_{RA_1})
 + \omega(a_i, RA_2) \cdot \pi(P_{RA_2}),
\label{eq cpt}
\end{equation}
where $P_{RA_j}$ is the objective probability of robot action $RA_j$, and $\pi(p)$ is expressed using the Prelec function \cite{akrenius2019entropic}:
\begin{equation}
\pi(p) = e^{-\beta (-\ln p)^{\alpha}}\ ,
\label{eqn_pi}
\end{equation}
with positive parameters $\alpha$ and $\beta$ controlling the curvature and elevation of the weighting function.

The probability that a participant chooses action $a_i$ can then be computed using the Noisy Rational Choice (softmax) model \cite{kwon2020humans}:
\begin{equation}
P(a_i) = \frac{e^{\lambda U^{\mathrm{C}}(a_i)}}{\displaystyle \sum_{\ell=1}^{2} e^{\lambda U^{\mathrm{C}}(a_\ell)}} ,
\label{eq softmax}
\end{equation}
where $\lambda>0$ reflects the degree of rationality and the utility $U^{\mathrm{C}}(a_i)$ is defined as in (\ref{eq cpt}). In the experimental setting with two possible actions, the participant’s compensation probability $P_2$ estimated using CPT, denoted as $P_2^{C}$, is given by
\begin{equation}
P_2^{C} := P(HA2) = \frac{1}{1 + e^{-\lambda \Delta U^{\mathrm{C}}}} \ ,
\label{eq 3}
\end{equation}
where $P_2^{C}$ is computed from experiments as in (\ref{eqn_P2}). Here $\Delta U^C$ is defined as
\bea
\Delta U^{\mathrm{C}} &=& U^{\mathrm{C}}(a_2) - U^{\mathrm{C}}(a_1) = -C + (G+V) \cdot \pi(P_{RA_2}) \nonumber\\
&=&-C + (G+V) \cdot \pi(P_R) 
\label{eq 4}
\eea
where $\pi(\cdot)$ is defined in (\ref{eqn_pi}).

We set $V = G = 1$ to normalize the scale of rewards and losses, which does not affect the qualitative predictions of the model because decisions depend on relative differences in subjective value. This also ensures that the remaining parameters are well-identified, with $C$ directly representing the subjective effort cost relative to one unit of reward or loss. With this the parameter vector 
\[
\bm{\theta} = [\alpha, \beta, C, \lambda]^T \in \mathbb{R}^4
\] 
can be estimated from the data collected in the experiments by minimizing the Negative Log-Likelihood (NLL) across all observed decision scenarios \cite{myung2003tutorial}. Here, the NLL is computed as
\bea
\text{NLL}(\bm{\theta}) = -\sum_{k=1}^{D} \left[ N_{2, k} \ln(P_{2, k}(\bm{\theta})) + N_{1, k} \ln(P_{1, k}(\bm{\theta})) \right],
\label{eq 6}
\eea
where $D$ is the number of perturbation probabilities ($D=5$ in our experiment). Here $N_{1,k}$ and $N_{2,k}$ are observed counts for HA1 and HA2, and $P_{1,k}(\bm{\theta})$ and $P_{2,k}(\bm{\theta})$ are the corresponding predicted probabilities under the CPT–softmax model.  

Parameter estimation for the CPT model was performed separately for each participant using their individual choice data. 
The optimization was conducted in Python using the SciPy optimization library, employing the L-BFGS-B algorithm \cite{zhu1997algorithm}. 
Bounds were imposed on all parameters to ensure numerical stability and to maintain them within their theoretically meaningful ranges. 
For each participant, the estimated parameters were those that maximized the likelihood of reproducing their observed choices across the five robot perturbation probabilities.

Finally, the parameters have the following interpretations:
\begin{itemize}
    \item $\alpha \in [0.5, 3] $: curvature of the probability weighting function (deviation from linear probability perception);
    \item $\beta \in [0.5, 5]$: elevation of the probability weighting function (overall probability distortion);
    \item $C \in [0.01, 5]$: subjective effort cost (shifts decision threshold horizontally);
    \item $\lambda \in [1, 30]$: choice determinism (higher $\lambda$ gives sharper transitions between HA1 and HA2).
\end{itemize}

\section{Results}

\subsection{Perturbation Responses}
\begin{figure}[htb!] 
\centering
\includegraphics[width=1.0\linewidth]{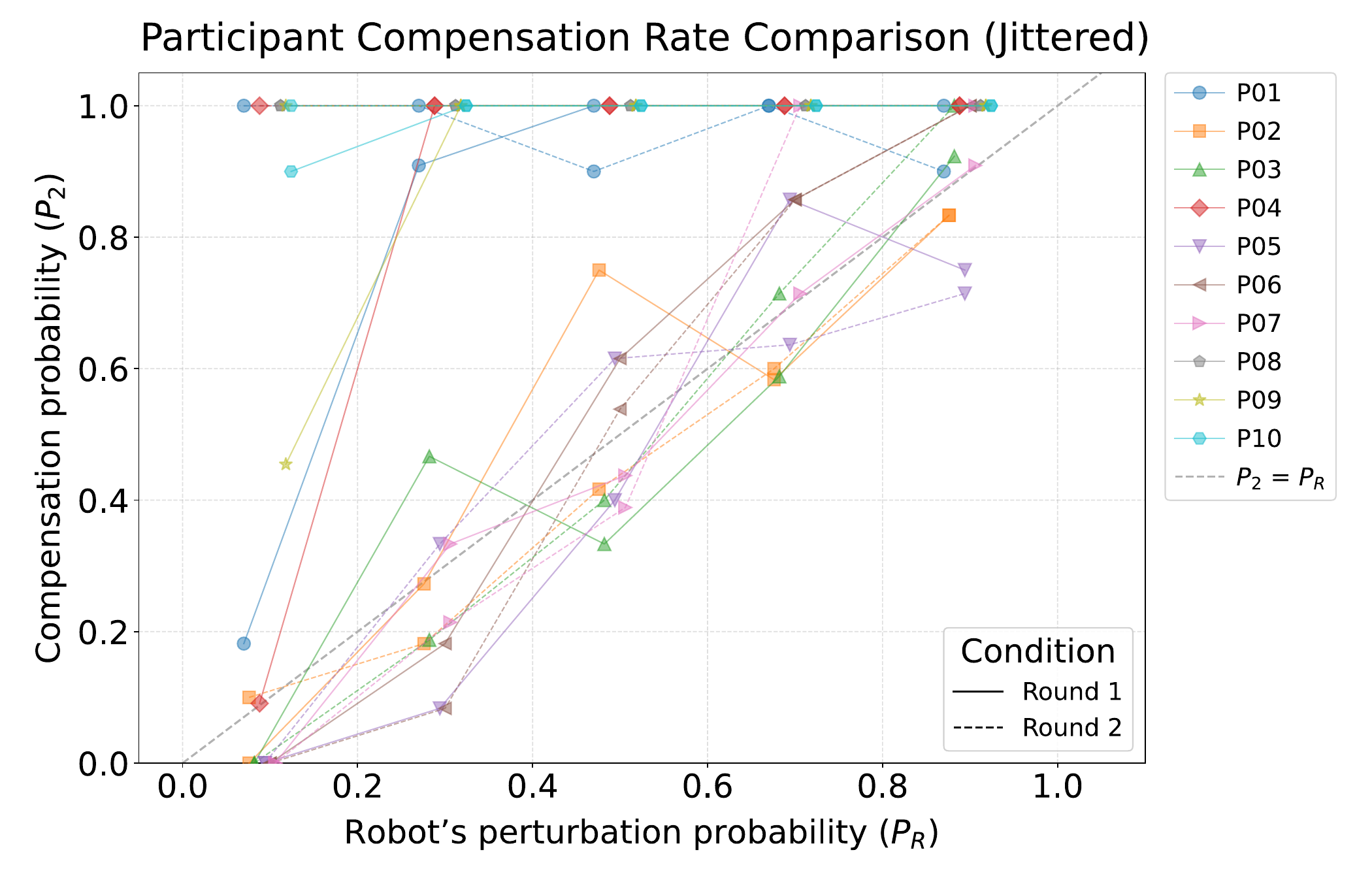}
\caption{Participant compensation probabilities across different robot perturbation probabilities. For each participant, results from the first round are shown with solid lines, and results from the second round with dashed lines. It is noted that some participants exhibited exactly the same action pattern in their second rounds, causing the corresponding dashed lines to coincide.}
\label{Fig. 3}
\end{figure}

The participant compensation probabilities across different robot perturbation probabilities are shown in Fig.~\ref{Fig. 3}. 
Two distinct action patterns can be identified from these results. 
The first pattern reflects a risk-averse, ``always-compensate'' strategy, observed in five participants, who consistently chose to compensate (HA2) regardless of the robot perturbation probability. 
The second pattern reflects a flexible, probability-dependent strategy, observed in the remaining five participants. 
For this group, the likelihood of choosing to compensate increased gradually with the perturbation probability, reflecting a trade-off between the expected reward of successful compensation and the subjective effort cost, rather than a fixed all-or-nothing behavior.

Within the ``trade-off'' group, compensation probabilities at the 10\% robot perturbation level typically appeared lower than the objective perturbation probability. 
This suggests that when the interruption probability is small, the subjective effort cost dominates, leading participants to under-react to low probability risks. 
Conversely, more than half of these participants exhibited an over-reactive tendency at high perturbation probabilities, reflecting strong loss aversion: participants were willing to incur additional effort to avoid an almost certain loss.

\subsection{Model Fitting}

\begin{figure}[h!]
\centering
\includegraphics[width=\linewidth]{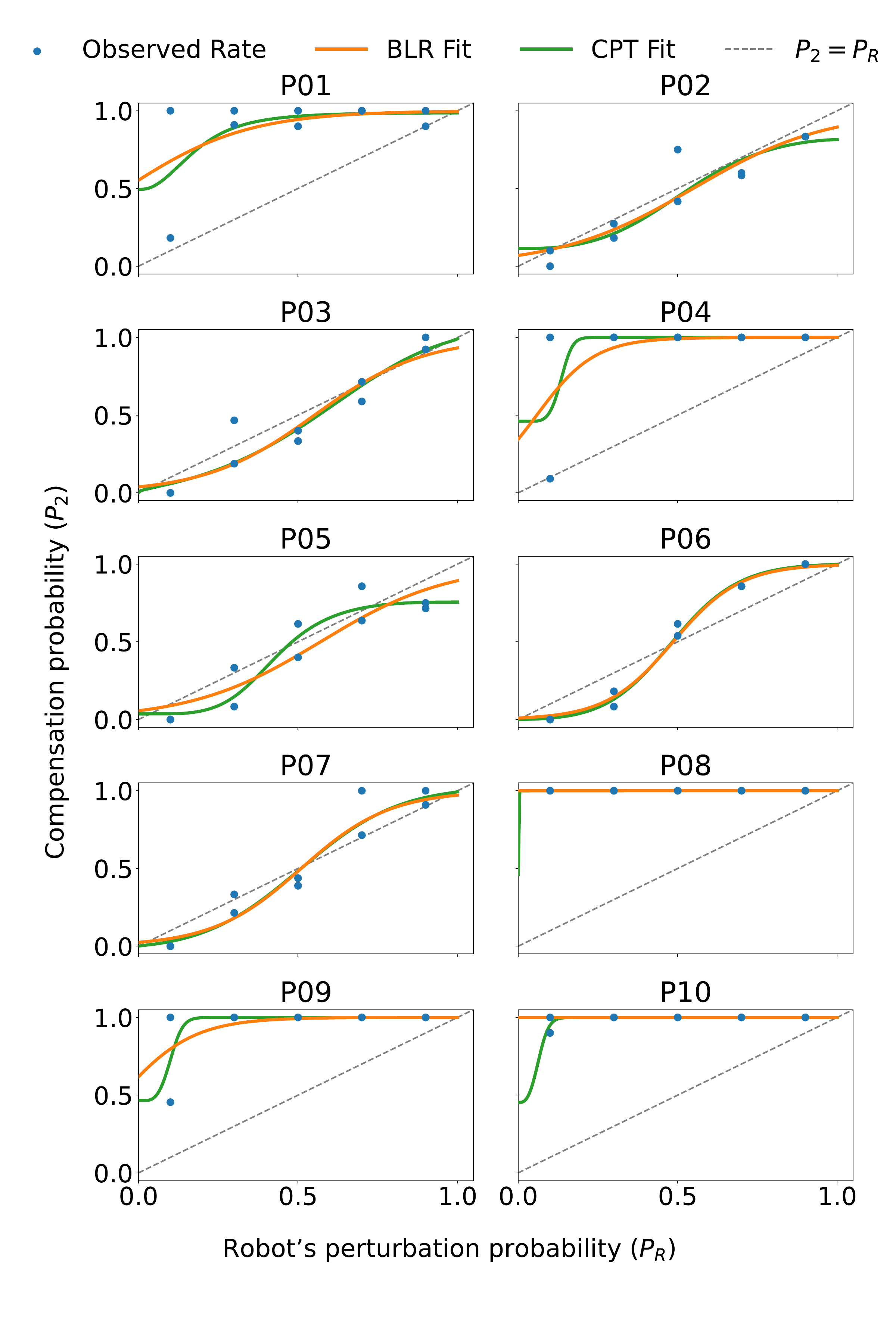}
\caption{Model fitting results to map perturbation probability to compensation probability. Blue dots denote the raw data from each round, while the red and green line show the BLR and CPT fits, respectively.}
\label{Fig. 6}
\end{figure}

Using two models: BLR (\ref{eq 1})  and CPT (\ref{eq 3}), we identified parameters of each models for each participant using their data. The fittings of both the BLR and CPT-based models for each participant are shown in Fig.~\ref{Fig. 6}. 
Both models were fitted in a data-driven manner: for each participant, one model (either BLR or CPT) was selected and its free parameters were estimated directly from that participant’s empirical choice frequencies across the five robot perturbation probabilities.

The fittings of both the BLR and CPT-based models for each participant is shown in Fig.\,\ref{Fig. 6}. The curves produced by the CPT–softmax model resemble those obtained from the BLR method. This demonstrates that both the BLR and CPT models have similar capability to capture the measure decision-making behaviors during pHRI. Here, it is noted that unlike purely statistical approaches such as BLR, CPT-based models provide interpretable insights into the psychological processes that underlie these behaviors, enabling a richer understanding of why different strategies emerge. A more detailed evaluation of each fit is provided in the text below.

\noindent \underline{BLR Fit}

Table~\ref{tab:fit_summary} summarizes the fitted BLR parameters and evaluation metrics for all participants. 
Two participants (P08 and P10) could not be fitted with a stable BLR model, as they consistently chose HA2 across all trials; 
without any variability in their choices, the model cannot reliably estimate the parameters. 
For the remaining participants, the fitted parameters vary across individuals, reflecting differences in decision-making strategies 
and indicating a degree of personalization in how participants weighed the trade-off between reward and effort cost.



\begin{table}[h!]
\centering
\caption{Individual Participant Bayesian Logistic Regression Summary}
\label{tab:fit_summary}
\begin{tabular}{l|cc|ccc}
\toprule
ID & $\bar{\beta}_0$ & $\bar{\beta}_1$ & $\Delta \bar{\beta}_0$ & $\Delta \bar{\beta}_1$ & $RMSE$ \\
\midrule
P01 & 0.21  & 5.26  & 2.09 & 6.43  & 0.20 \\
P02 & -2.61 & 4.77  & 1.99 & 3.42  & 0.11 \\
P03 & -3.23 & 5.85  & 2.05 & 3.62  & 0.11 \\
P04 & -0.61 & 10.83 & 2.52 & 11.27 & 0.21 \\
P05 & -2.79 & 4.92  & 1.99 & 3.27  & 0.12 \\
P06 & -4.58 & 9.46  & 3.17 & 6.03  & 0.04 \\
P07 & -3.70 & 7.26  & 2.52 & 4.71  & 0.09 \\
P08 & 10.00 & 7.26  & 2.52 & 9.96  & 0.00 \\
P09 & 0.50  & 8.87  & 2.64 & 11.55 & 0.13 \\
P10 & 10.00 & 8.87  & 2.64 & 8.17  & 0.03 \\
\bottomrule
\end{tabular}
\end{table}



The reliability of the BLR parameter estimates was quantified using both the root mean squared error (RMSE) of the model fit 
and the 95\% confidence intervals ($\Delta \bar{\beta}_0$ and $\Delta \bar{\beta}_1$), where wider intervals indicate greater uncertainty 
in the parameter estimates. The interplay between RMSE and confidence interval width provides insight into participants' underlying decision strategies.

Participants in the ``always-compensate'' cluster (e.g., P08 and P10) exhibited a deterministic strategy, consistently choosing HA2. 
This produced minimal prediction errors (low RMSE), but the lack of behavioral variability resulted in wide confidence intervals for the fitted parameters 
(high $\Delta \bar{\beta}$), as the ``step'' response conveys little information about the slope beyond its steepness.
In contrast, participants in the ``trade-off'' cluster displayed a probabilistic, graded strategy. 
Although this introduced slightly higher prediction errors (higher RMSE), the variability in their responses across perturbation probabilities allowed the model 
to estimate the logistic curve parameters with high precision (low $\Delta \bar{\beta}$).

Therefore, wide confidence intervals in the ``always-compensate'' group should not be interpreted as a model failure, but rather as a signature of a highly decisive behavioral strategy.

\begin{figure}[h!]
\centering
\includegraphics[width=\linewidth]{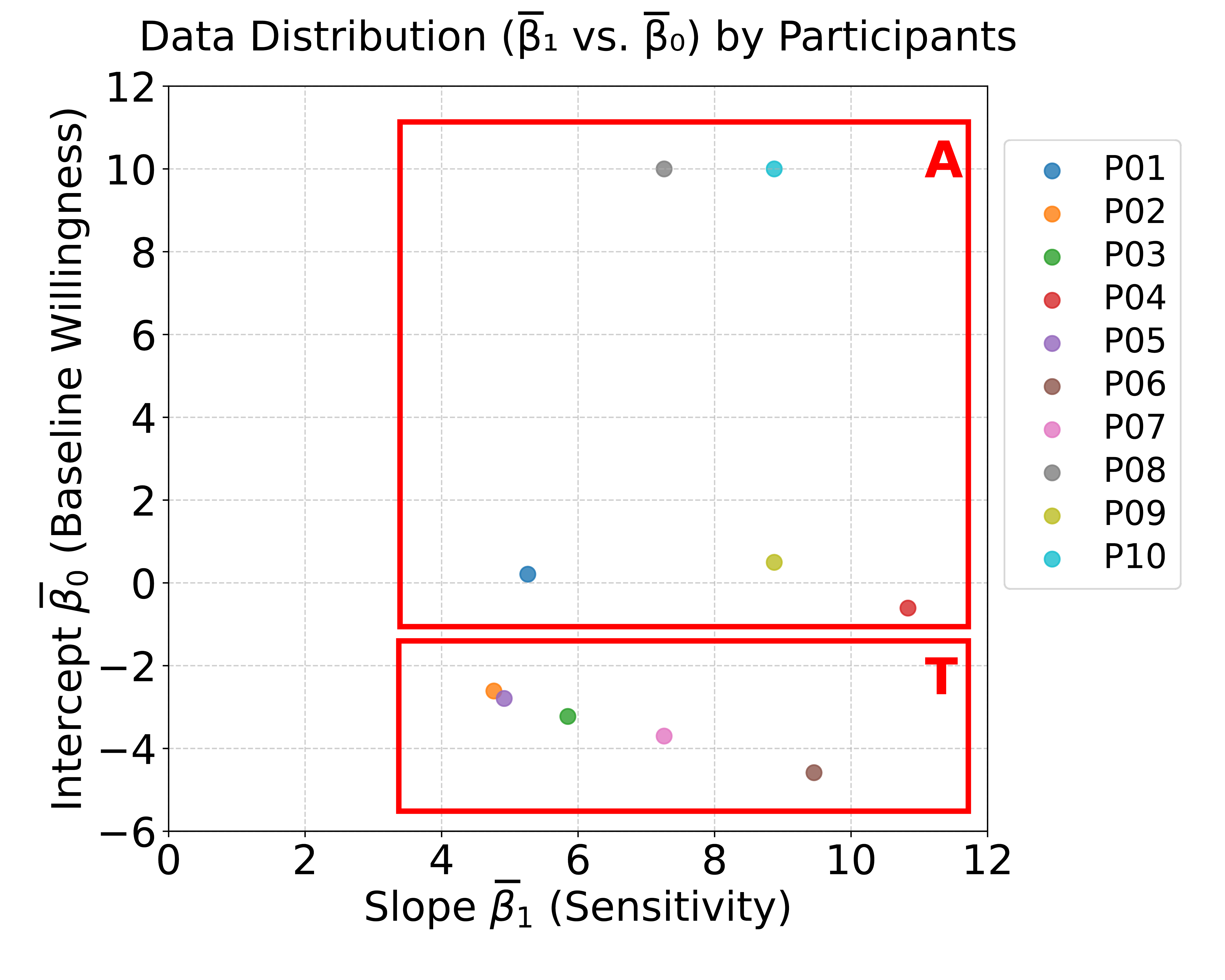}
\caption{Distribution of the logistic regression fitting results for all participants. The red boxes highlight the two distinct clusters: A-"always-compensate" strategy; T-"trade-off" strategy.}
\label{Fig. 5}
\end{figure}

The distributions of the fitted BLR parameters $\bar{\beta}_0$ and $\bar{\beta}_1$ are shown in Fig.~\ref{Fig. 5}, 
and two distinct clusters can be identified, consistent with the patterns observed in Fig.~\ref{Fig. 3}. 
The first cluster is characterised by $\bar{\beta}_0$ values close to zero or positive, reflecting a highly risk-averse, ``always-compensate'' strategy. 
In contrast, the second cluster exhibits $\bar{\beta}_0$ values in the range $[-2, -5]$, corresponding to a more step-like, cost–reward trade-off strategy. 

Within this second cluster, the slope parameter $\bar{\beta}_1$ captures further heterogeneity in participants' responses to changes in perturbation probability. 
Participants with lower $\bar{\beta}_1$ values (e.g., P02, P03) show a more gradual sensitivity, implying a smoother transition in their decision-making. 
Participants with higher $\bar{\beta}_1$ values (e.g., P06) demonstrate a sharper, more step-like sensitivity, indicating a decisive switch in strategy once a particular probability threshold is reached.

\noindent \underline{CPT Fit}

Table~\ref{tab:cpt_params} presents the fitted parameters obtained using the CPT-softmax behavioral model, 
with parameter bounds set to $\alpha \in [0.5, 3]$, $\beta \in [0.5, 5]$, $C \in [0.01, 5]$, and $\lambda \in [1, 30]$. 
For participants adopting the ``always-compensate'' strategy, the fitted subjective effort cost $C$ reached the lower bound of the fitting range. 
This suggests that for these participants, the additional physical effort required for compensation was perceived as relatively small compared to the benefit of ensuring trial success. 

\begin{figure}[h!]
\centering
\includegraphics[width=\linewidth]{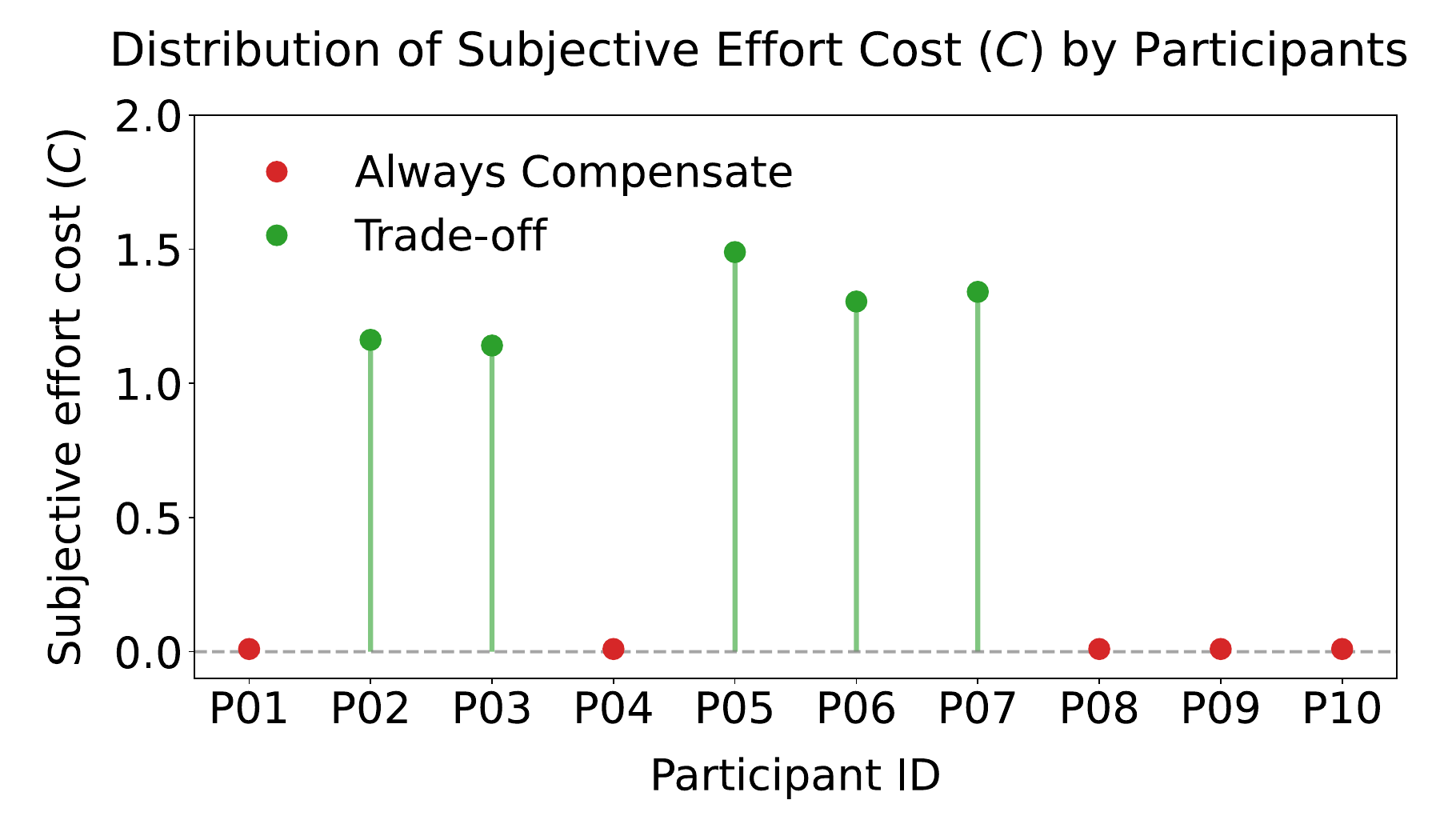}
\caption{Distribution of the participant subjective effort cost $C$. Red dots denote participants who always compensated while green denotes participants in the trade-off cluster.}
\label{Fig. cpt_c}
\end{figure}

Based on the fitted $C$ values (Fig.~\ref{Fig. cpt_c}), two distinct clusters are evident. 
The first cluster, with small $C$ values, corresponds to participants who consistently adopted an ``always-compensate'' strategy. 
The second cluster, with $C$ values larger than 1, corresponds to participants employing a graded ``trade-off'' strategy, 
where the decision to compensate depends on the balance between expected reward and perceived effort cost.

\begin{table}
\centering
\caption{ESTIMATED CPT-SOFTMAX MODEL PARAMETERS SUMMARY}
\label{tab:cpt_params}
\begin{tabular}{lcccc}
\toprule
Subject & $\alpha$ & $\beta$ & $C$ & $\lambda$ \\
\midrule
    P01 & 1.86 & 0.50 & 0.01 & 2.15 \\
    P02 & 1.61 & 1.17 & 1.16 & 1.76 \\
    P03 & 0.50 & 0.74 & 1.14 & 5.63 \\
    P04 & 2.72 & 0.50 & 0.01 & 15.80 \\
    P05 & 2.12 & 0.50 & 1.55 & 2.68 \\
    P06 & 0.67 & 0.53 & 1.30 & 9.21 \\
    P07 & 0.57 & 0.50 & 1.34 & 7.79 \\
    P08 & 0.50 & 0.50 & 0.01 & 15.65 \\
    P09 & 2.23 & 0.50 & 0.01 & 14.08 \\
    P10 & 1.91 & 0.51 & 0.01 & 18.78 \\
\bottomrule
\end{tabular}
\end{table}

\section{Discussion}
\iftrue
This pilot study investigated how humans make decisions when a robot can perform one of two actions (assistive and perturbing), each applied with a specified probability during real-time pHRI, where the effort required to complete the task and potential losses were implicit. The results revealed two distinct strategies: some participants adapted their behavior according to the probability of perturbing, while others maintained a consistent strategy across all blocks. Among participants who adapted their behavior, there was a slight distortion in the perception of extreme probabilities, broadly consistent with predictions from models such as CPT.

The presence of clustered behavior, together with the systematic deviations observed among participants who engaged in a genuine trade-off between effort and task success, reflects the personalized nature of human decision-making in pHRI. For these participants, the trade-off involved deciding whether to invest additional effort through compensation or to conserve effort and accept a greater risk of failing to reach the target if the robot produced a perturbation. Although some variability could be interpreted as noise within human decision-making, as suggested by noisy rational models \cite{kwon2020humans}, the emergence of distinct behavioral clusters indicates a deeper divergence in how individuals interpret the same probabilistic cues. These differences can instead be viewed through the lens of CPT, particularly the idea that decisions are evaluated relative to an internal reference point. Variation in these reference points, as observed through the relative cost parameter ($C$), offers a natural explanation for the observed behavioral clusters. Participants who adopted a constant strategy, for example, offered justifications during brief post-experiment discussions such as ``I trust myself more than the robot, so I always compensate in advance'' or ``I only care about succeeding; the extra effort is not a concern,'' revealing stable internal biases rather than random fluctuations. This suggests that CPT-based models for pHRI should incorporate such biases explicitly, including factors related to trust, subjective cost sensitivity, and individual reference points.

Interestingly, participants tended to underestimate low-probability perturbations (10\%) and overestimated high-probability perturbations (90\%), suggesting asymmetric sensitivity to uncertainty across probability levels. This warping of probability perception has been observed in multiple studies of human behavior \cite{kwon2020humans, smith2025if, prashanth2016cumulative}, but it has received little attention in a pHRI context. Future pHRI algorithms, including those based on stochastic optimal control and game-theoretic frameworks, should consider modifying expected value computations to account for these systematic biases in human probability perception. However, such modifications are likely to be context-dependent, highlighting the need to collect sufficient empirical data to develop accurate models.

It is worth noting that the observed results could be captured both by classic Bayesian logistic regression, which is entirely data-driven and does not provide insight into the behavioral mechanisms underlying human decision-making, and by a CPT framework, which explicitly accounts for behavior. Although our results suggest that CPT can represent many of the features observed in our pHRI experiment, we cannot claim a direct application of CPT in this context because the small dataset led to some numerical instability in parameter fitting. In particular, varying the initial settings caused significant fluctuations in the fitted $\alpha$ and $\beta$ results, which limiting confidence in the estimated parameters. Furthermore, since both models were able to explain the observed behavior, our findings do not necessarily imply that CPT is the optimal model for pHRI. Future studies should consider modifications to the experimental design and constraints that produce divergent predictions across models, enabling clearer validation. Here, it is also notable that some participants anecdotally reported basing their decisions on previous trials. This form of sequential or Bayesian reasoning is not naturally captured by existing CPT models, but it may reflect an important factor to consider in future adaptations of CPT for pHRI.

We acknowledge that the parameter identification techniques for the BLR-based and CPT-based models in this study differ. As this is a preliminary study, our primary focus is to demonstrate that CPT-based modeling provides a more intuitive understanding of participant decision-making. With the limited number of subjects, both parameter identification techniques are constrained and may not precisely estimate all parameters. In future work, the experimental protocol will be carefully redesigned and conducted with a larger participant cohort to ensure robust convergence of the parameter estimation procedures.

There are a number of other limitations in our study design that constrain the potential generalizability of these findings. One critical factor is the use of blockwise randomization with either increasing or decreasing perturbation probabilities. Although the ordering was balanced across participants, there were signs of learning effects; for example, some participants in the constant-strategy cluster only exhibited different behavior in their first block. Additionally, the blockwise design may have encouraged participants to perceive patterns within a block, rather than treating each reach as an independent event. Both of these issues could be addressed in future studies by implementing trial-by-trial randomization.

It is also worth noting that our experiment constrained both human and robot actions to a small finite set and that the task was primarily feedforward in nature. While this approach is consistent with many previous applications of CPT \cite{o2019asymmetric, kwon2020humans}, it does not fully capture typical pHRI scenarios, where tasks often involve feedback and human actions are multi-dimensional and potentially continuous. Future studies should consider how to extend these models to accommodate a broader range of possibilities. One starting point could be to design tasks that are more physically demanding and to explore scenarios in which success and failure are not strictly binary outcomes.

\else

\begin{itemize}
    \item Summary of results
    \begin{itemize}
        \item Study was to investigate how humans respond to uncertainty during pHRI.
        \item The findings show a few important elements: distinct human behaviors fitting in different clusters; differences relative to true probability; CPT-based modeling appears able to capture some of these features.
    \end{itemize}
    \item Clusters and reference points {\bf Adding implicit cost}
    \begin{itemize}
        \item Two distinct behaviors are seen (the two clusters).
        \item Within the clusters there is also distinct behavior (especially for those that showed a trade-off behavior)
        \item In CPT-based modeling this is related to reference points which can capture many of these features.
        \item Use reference points to explain the two observed behaviors.
        \item The presence of the clusters is though something that CPT would not necessarily predict. This may be indicative internal factors for the individual participants.
        \item Understanding that is a point for future work.
    \end{itemize}
    \item Understanding of probability
    \begin{itemize}
        \item Results suggest that participants often underestimated the 10\% probability (and there was a trend to overestimate large probabilities).
        \item This finding is not typically considered within pHRI design but has been observed across multiple other studies of human behavior \cite{kwon2020humans}.
        \item Future pHRI therefore should be at least revised to reflect human understanding of probability. 
        \item That understanding is context dependent and therefore there is a need to collect sufficient data with which to build the necessary models.
    \end{itemize}
    \item CPT for future use in pHRI
    \begin{itemize}
        \item The use of CPT modeling was able to capture important features of the model
        \item However, its direct use is limited in this study given the lack of data and with the data collected we cannot necessarily prove that CPT is better than other existing models for modeling human behavior in response to uncertainty.
        \item It is also noted that anecdotally participants reported consideration of prior events as influencing their behavior.
        \item This Bayesian approach is not naturally captured in existing CPT which is predominately modeling macro level events.
        \item For application in pHRI this as an implicit factor needs to be accounted for, where our belief is that CPT is more suited to such an accounting.
    \end{itemize}
    \item Limitations of order
    \begin{itemize}
        \item There are a number of limitations in our study design that need to be acknowledged.
        \item Ordering effect and blockwise nature may have influenced some of the outcomes. 
        \item Here, for example most of the members of the "always compensate" cluster who show deviation from their typical behavior specifically did so in their first block indicating a learning effect.
        \item Additionally, the blockwise nature may have meant that participants were more prone to ``counting" behaviors rather than truly responding on an independent level.
        \item Future data collection should expand on our results to evaluate such effects
    \end{itemize}
    \item Realism with respect to pHRI
    \begin{itemize}
        \item Its also worth noting that the cost considered and set of possible options are simplified relative to true pHRI
        \item Here, the task likely needs to be more difficult to account for the ``physical" component. Moreover the idea of success and failure likely need to be clearer.
        \item Similar to this, pHRI differs from other applications of CPT in that there is typically an infinite set of possible actions by both the robot and human.
        \item How to account for this broader set of possibilities needs to be considered in future.
    \end{itemize}
\end{itemize}

\fi

\section{Conclusion}

This pilot study investigated human decision-making in scenarios where the robot could execute one of two actions, assistive or perturbing, each occurring with a predefined probability during real-time pHRI. Participants exhibited two main behavioral patterns. Some adjusted their responses according to the likelihood of perturbation, while others maintained a consistent, risk-averse strategy across trials. Among those who adapted their behavior, systematic distortions in the perception of extreme probabilities were observed, broadly consistent with predictions from cognitive frameworks such as CPT. 

These findings underscore the variability and personalization of human decision-making in pHRI and suggest that CPT may provide a useful framework for characterizing these behaviors. Due to the limited sample size and experimental constraints, definitive conclusions cannot yet be drawn. Future work should implement fully randomized trial sequences, expand the participant pool, and explore a broader range of robot actions. Such efforts will improve model validation and support the development of adaptive robot controllers that account for personalized human decision-making, enhancing collaboration, safety, and performance in shared human–robot environments.

\section*{Acknowledgment}
The authors acknowledge funding received by the Australian Research Council (grant number DP240100938) that supported this research.

\bibliographystyle{IEEEtran}
\bibliography{Reference}

\end{document}